\documentclass{article}

\usepackage{microtype}
\usepackage{graphicx}
\usepackage{subfigure}
\usepackage{amsmath}
\usepackage{amssymb}
\usepackage{booktabs} 
\usepackage{capt-of}
\usepackage[symbol]{footmisc}

\usepackage{hyperref}

\usepackage{algpseudocode}
\usepackage{algorithmicx}

\usepackage[accepted]{icml2019}

\icmltitlerunning{GANchors: Realistic Image Perturbation Distributions
                for Anchors Using Generative Models}
\begin{document}

\twocolumn[
\icmltitle{GANchors: Realistic Image Perturbation Distributions\\
                for Anchors Using Generative Models}



\icmlsetsymbol{equal}{*}

\begin{icmlauthorlist}
\icmlauthor{Kurtis Evan David}{ed1}
\icmlauthor{Harrison Keane}{ed2}
\icmlauthor{Jun Min Noh}{ed2}
\end{icmlauthorlist}

\icmlaffiliation{ed1}{Department of Computer Science, University of Texas at Austin}

\icmlaffiliation{ed2}{Department of Electrical and Computer Engineering, University of Texas at Austin}

\icmlcorrespondingauthor{kurtis.e.david@gmail.com}

\vskip 0.25in
]



\printAffiliationsAndNotice{}  

\begin{abstract}
We extend and improve the work of Model Agnostic Anchors for explanations on image classification through the use of generative adversarial networks (GANs). Using GANs, we generate samples from a more realistic perturbation distribution, by optimizing under a lower dimensional latent space. This increases the trust in an explanation, as results now come from images that are more likely to be found in the original training set of a classifier, rather than an overlay of random images. A large drawback to our method is the computational complexity of sampling through optimization; to address this, we implement more efficient algorithms, including a diverse encoder. Lastly, we share results from the MNIST and CelebA datasets, and note that our explanations can lead to smaller and higher precision anchors.\footnote[1]{The associated code and datasets can be found at: \href{https://github.com/kurtisdavid/ImageAnchors}{https://github.com/kurtisdavid/ImageAnchors}}
\end{abstract}

\section{Introduction}
As use of machine learning models becomes increasingly regulated, explainability of models similarly becomes a pressing issue. For example, the EU's 2018 GDPR regulation stipulates that all individuals have a right to an explanation of the decisions reached using their data. Explainability of models can be approximated by global interpretability of sets of predictions, or local interpretation of individual predictions. \cite{DBLP:conf/aaai/Ribeiro0G18} proposed Model Agnostic Anchors as a method of achieving local interpretability. Their work formally defines an anchor $A$ by the following:

\begin{equation}
    \mathbb{E}_{\mathcal{D}(z|A)}[1_{f(x)=f(z)}] \geq \tau, A \subseteq x
\end{equation}

i.e. a subset of features in a datapoint $x$ where, if the other features are perturbed in some distribution $\mathcal{D}$, then the prediction of the model remains the same on the new sampled data $z$. The threshold $\tau$ is a desired confidence, and for their tests, they use $\tau = 0.95$. However, since this expectation is intractable, so they provide a probabilistic definition for (1), given that (1) defines $prec(A)$:
\begin{equation}
    P\bigl(prec(A) \geq \tau\bigr) \geq 1-\delta
\end{equation}

i.e. with $\delta$ high confidence, $prec(A) \geq \tau$. So long as this is achieved, then we can consider $A$ as an anchor. However, to find the "best" anchor, they introduce their optimization on the coverage of an Anchor, where $cov(A) = \mathbb{E}[A \subseteq z]$:
\begin{equation}
    \max_{A\ s.t.\ P(prec(A) \geq \tau) \geq 1-\delta} cov(A)
\end{equation}

To find the anchors, they approximate the highest precision calculation by formulating the search as a \textit{multi-armed bandit} problem. They utilize the KL-LUCB algorithm to solve this. Furthermore, because this is a purely greedy approach, to obtain a better possible solution for (3), they also implement beam search as a way to increase exploration of the solution space. 

\begin{figure}[t]
\vskip 0.2in
\begin{center}
\centerline{\includegraphics[width=\columnwidth]{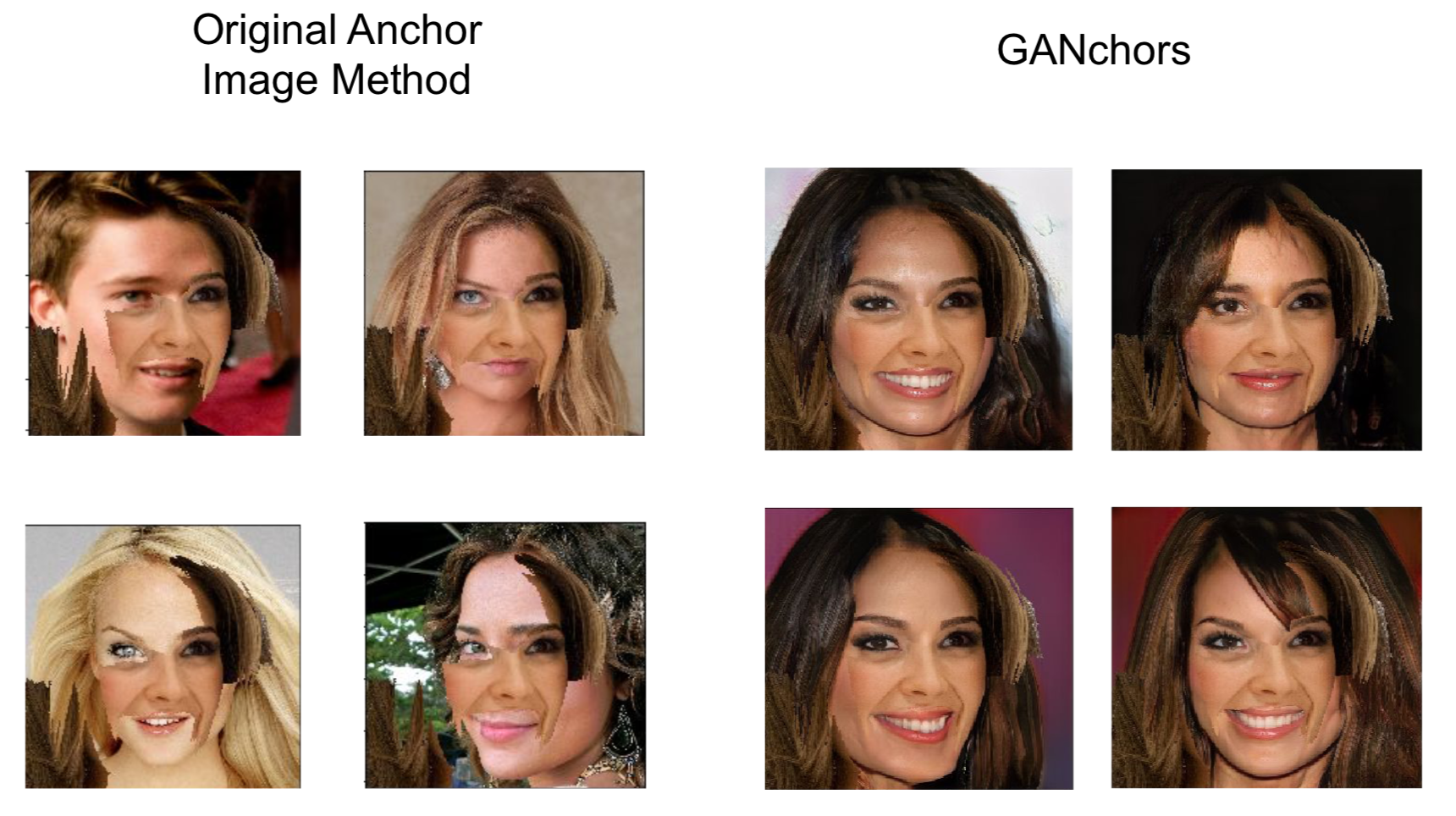}}
\caption{Comparison of samples generated using the original method with the samples generated with GANchors. Our samples are a more faithful representation of the true underlying data distribution. }
\label{icml-historical}
\end{center}
\vskip -0.2in
\end{figure}

\begin{figure}[ht]
\vskip 0.2in
\begin{center}
\centerline{\includegraphics[width=\columnwidth]{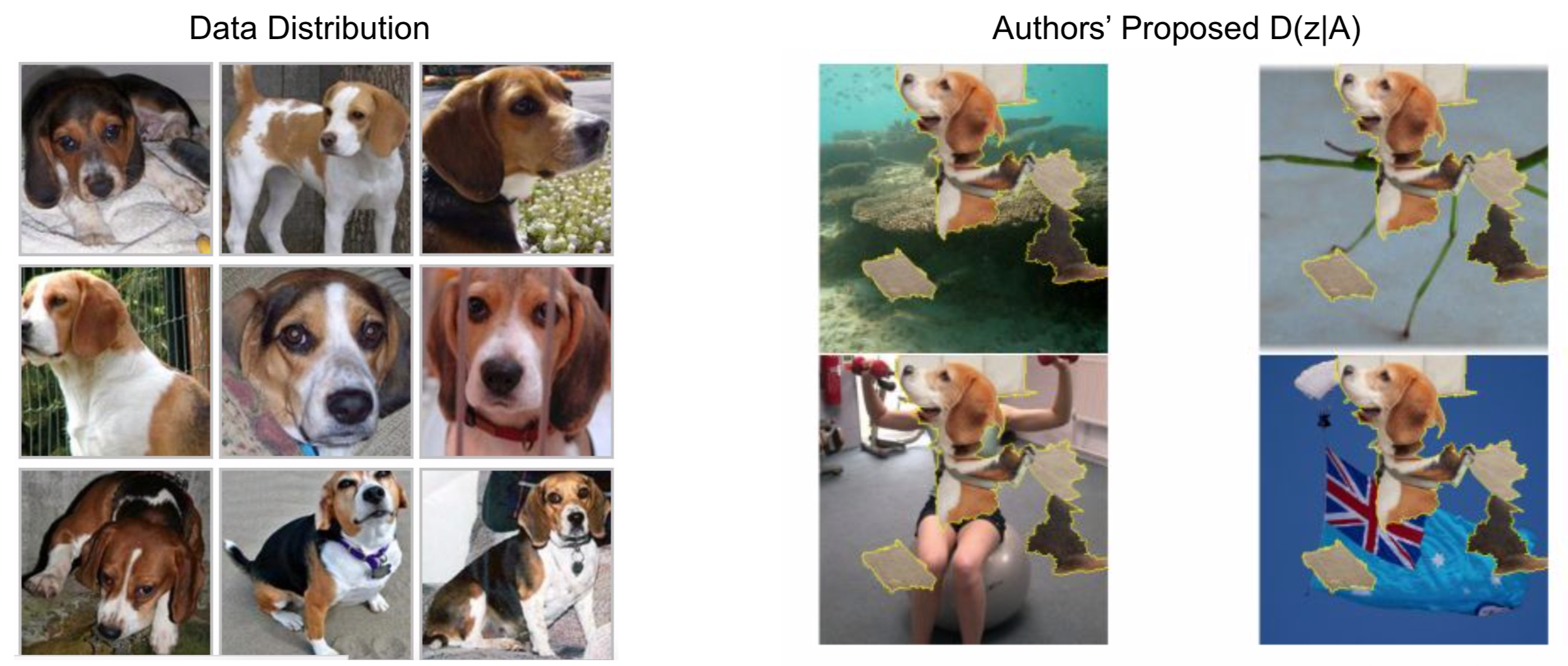}}
\caption{Original Anchor Method does not create images that are representative of the underlying distribution the model was trained for.}
\label{icml-historical}
\end{center}
\vskip -0.2in
\end{figure}

\section{Motivation and Contributions}

An important part of Anchors \cite{DBLP:conf/aaai/Ribeiro0G18} is the definition of $D(z|A)$, representing a perturbation distribution that lies in the original training distribution. For domains such as text, this was achieved via replacing non anchored words in sentences with new words of the same parts of speech and word embeddings of the original words. However, the authors did not use a realistic $D(z|A)$ for the image domain examples, instead producing them by overlaying segmentations onto random test images as can be seen in Figure 2.

The significant contributions of this paper are as follows:

\begin{enumerate}
    \item We improve anchors for images by defining a more realistic $D(z|A)$ by performing stochastic gradient descent on a trained generative network.
    \item We propose several methods of reducing computational complexity of optimization, including a \textit{diverse encoder} to obtain initial projections to the latent space.
\end{enumerate}

Our work improves anchors for images by generating samples from a perturbation distribution that is more consistent with the authors' original theory. Although the random image and the potential anchors come from the same underlying distribution, the combination of them both result in images that are not represented in the data distribution and are more representative of being an outlier. We do not believe an explanation based on samples generated from such a distribution are trustworthy. Results from this type of analysis can result from unexplained feature interactions, depending on the sensitivity of the desired classification model. To amend this, we propose using Generative Adversarial Networks \citep{2014arXiv1406.2661G}, trained to generate new samples that come from the original training distribution.

\section{Proposed Method and Related Work}

\subsection{Generative Networks}
Generative Models are a class of models that attempt to learn the distribution of the data and generate samples from a representative distribution. While this class includes various models such as bayesian networks and Markov Random Fields, for this work we focus on generative adversarial networks, or GANs. GANs typically consist of two networks: a generator and a discriminator that compete in a minimax objective. The goal of the discriminator is to classify images as real or fake, while the generator’s goal is the fool the discriminator by generating images indistinguishable from the real distribution. To sample from a generator, the generator is trained to map random noise vectors $z$ in a lower dimensional latent space to the domain of the target distribution. If successfully trained, the generator is able to create new and diverse samples that mimic the original dataset. We will refer to this as the manifold of the GAN for the rest of the paper.

This ability to model underlying distributions combined with recent advancements in the field make GANs suitable for sampling perturbations of data points for the Anchor algorithm. In addition, researchers have been moving towards sharing trained models, making it easier to access high quality reconstructions without having to train complex networks ourselves. For testing, we explored several pretrained GANs, including DCGAN \citep{DBLP:journals/corr/RadfordMC15}, BigGAN \citep{brock2018large}, and ProGAN \citep{karras2018progressive}.

\subsection{Image In-painting}
One key observation to be made with their current image framework is their use of superpixels as explainable features made from a segmentation map. This map contains labels for each pixel, thus to extract a certain segment $i$ from the image $x$, one can apply an element wise product with a binary matrix $A$ to extract a masked image $\hat{x} = A \circ x$ (we simplify this to $Ax$ for the rest of the paper). Thus, our task is closely related to solving the following optimization:
\begin{equation}
    \min_y || \hat{x} - Ay || 
\end{equation}

The result of equation (1) would reconstruct an image $y$ that would match the masked areas of the image the most, under a norm such as $L2$. Because $A$ describes a binary segmentation map, this problem can be viewed as image inpainting, i.e. the task of filling in holes of a masked image. There are multiple applications for image inpainting due to its versatility, but the most common use for it is to remove undesired part of an image and replace it with realistic pixels that blends in with the rest of the image. However, our purpose of inpainting is not to just fill in with plausible pixels, but also to generate images that are realistic in terms of context. Another challenge that with inpainting is that the superpixels found during the Anchor algorithm are relatively small compared to the rest of the image. This means that the hole that needs to be filled in is larger than the part of the image that the inpainting method is optimizing over.

One state-of-the-art inpainting method, PatchMatch, fills in masked holes with smooth results, but uses image statistics of the unmasked image to fill in the missing portion \citep{Barnes:2009:PAR}. This prevents the method from using a semantically-aware approach and often fills in the holes with undesired parts of superpixels. Additional limitations of other methods include strong assumptions of masked holes. For example, Pathak et al. assumes that the masked hole is a rectangle with dimensions of 64 $\times$ 64 that is centered in the middle of the image \citep{pathakCVPR16context}. This would be unfeasible for our research since random segmentation can return scattered and irregular superpixels. Finally, Partial Convolution \citep{Liu_2018_ECCV} is an inpainting method proposed by NVIDIA that uses deep learning to fill irregular holes by learning semantic priors and hidden representations. However, there are still limitations of this model in that it is deterministic and there is not a way to properly generate multiple realistic perturbations given an anchor.

\subsection{Compressed Sensing with Generative Models}

Image inpainting is one problem contained in the topic of Compressed Sensing. The goal is to reconstruct original measurements $x^*$ given lossy measurements $\hat{x}$ defined by a measurement matrix $A$.
\[
    \hat{x} = Ax^* + \eta
\]
This equation is actually overdetermined, since $\hat{x}$ actually lies in a lower dimensional subspace. To solve this and create natural solutions, a sparse prior is induced on possible reconstructions $y$. We can reframe Equation (4) to:
\begin{equation}
    \min_y || \hat{x} - Ay || + \lambda ||y||_1
\end{equation}

However, this can still be computationally heavy if the domain is in the image space. Bora et al. suggested alternative to (5) by restricting possible reconstructions $y$ on the manifold of a trained GAN \citep{pmlr-v70-bora17a}. This can be found in Equation (3). This enforces a natural image prior based off of the quality of the GAN, and reduces the optimization to a significantly lower-dimensional latent space. One of their applications was image inpainting, and we believe that there is promise in utilizing their framework due to the increasing capabilities of these generative networks, as well as being able to mimic the original training set -- the exact problem we are trying to solve.

\begin{equation}
    \min_z || \hat{x} - AG(z) || 
\end{equation}

\subsection{Realistic Image Sampling}

Under the optimization of (6), we observe that this is highly non-convex, due to the complex structure of generative networks. We can use this observation to apply a simple sampling algorithm; depending on the initialization of the latent vector $z \in \mathbb{R}^d$, we may obtain very different reconstructions from $G(z)$. Additionally, because it is unfeasible to repeatedly optimize (2) to small values, we introduce a reconstruction threshold $\xi$. We then define valid reconstructions $y$ from the GAN that contain an anchor $A$ as:
\begin{equation}
    y = G(z)\ s.t.\ ||\hat{x} - Ay|| < \xi
\end{equation}

Reconstructions $y$ can now be found by running gradient descent. However, to use these reconstructions in the Anchor algorithm, we must guarantee that $\hat{x}$ is within the image. Thus, the final reconstruction uses the generated image to fill in the holes of $\hat{x}$ found in (4) as well as Algorithm 1. Examples compared to original random stitching method can be seen in Figure 4.
\begin{equation}
    y = (1-A)G(z) + \hat{x}\ s.t.\ ||\hat{x} - AG(z)|| < \xi
\end{equation}

\begin{algorithm}
  \caption{Sample $y\sim D(\cdot|A)$
    \label{alg:packed-dna-hamming}}
  \begin{algorithmic}[1]
    \Require{Trained GAN $G$, Threshold $\xi$, Anchor $A$, Target $\hat{x}$, Learning rate $\alpha$}
    \Function{Sample}{$G, \xi, A, \hat{x}, \alpha$}
      \State $z \gets \mathcal{N}(0,1)$
      \While{$||\hat{x} - AG(z)|| > \xi$}
        \State $z \gets z - \alpha \nabla ||\hat{x} - AG(z)||$
      \EndWhile
      \State $y \gets (1-A)G(z) + \hat{x}$
      \State \Return{$y$}
    \EndFunction
  \end{algorithmic}
\end{algorithm}

Following the midterm project, we realized that our original algorithm is biased towards generating samples immediately below the threshold $\xi$. There exist other possible images that could have lower anchor reconstructions, so we wanted to extend our work to include these possibilities. To implement this, we pass different thresholds $< \xi$ into our reconstruction algorithm, sampled according to a threshold distribution. However, this can become computationally inefficient if much smaller thresholds are consistently used.

Thus, we first apply a probabilistic prior on possible reconstructions. Specifically, we contend that the densities of these reconstructions decreases as $\xi$ decreases. This is because at each possible level, there will be fewer local minima that can reach small reconstruction errors on the manifold. Figure 3 visualizes this concept. To incorporate this prior, we can define a decreasing probability distribution to sample other thresholds $\xi' < \xi$ where $\xi$ is the original max threshold. Algorithm 2 provides an example we use in our project on how to do this. We are essentially collecting points centered at $\xi$ and throwing away any invalid thresholds. We also use a small standard deviation to mimic an exponential decrease of densities. Future work can make this more accurate or learn the true distribution of our algorithm.

\begin{figure}[h]
\begin{center}
\centerline{\includegraphics[width=\columnwidth]{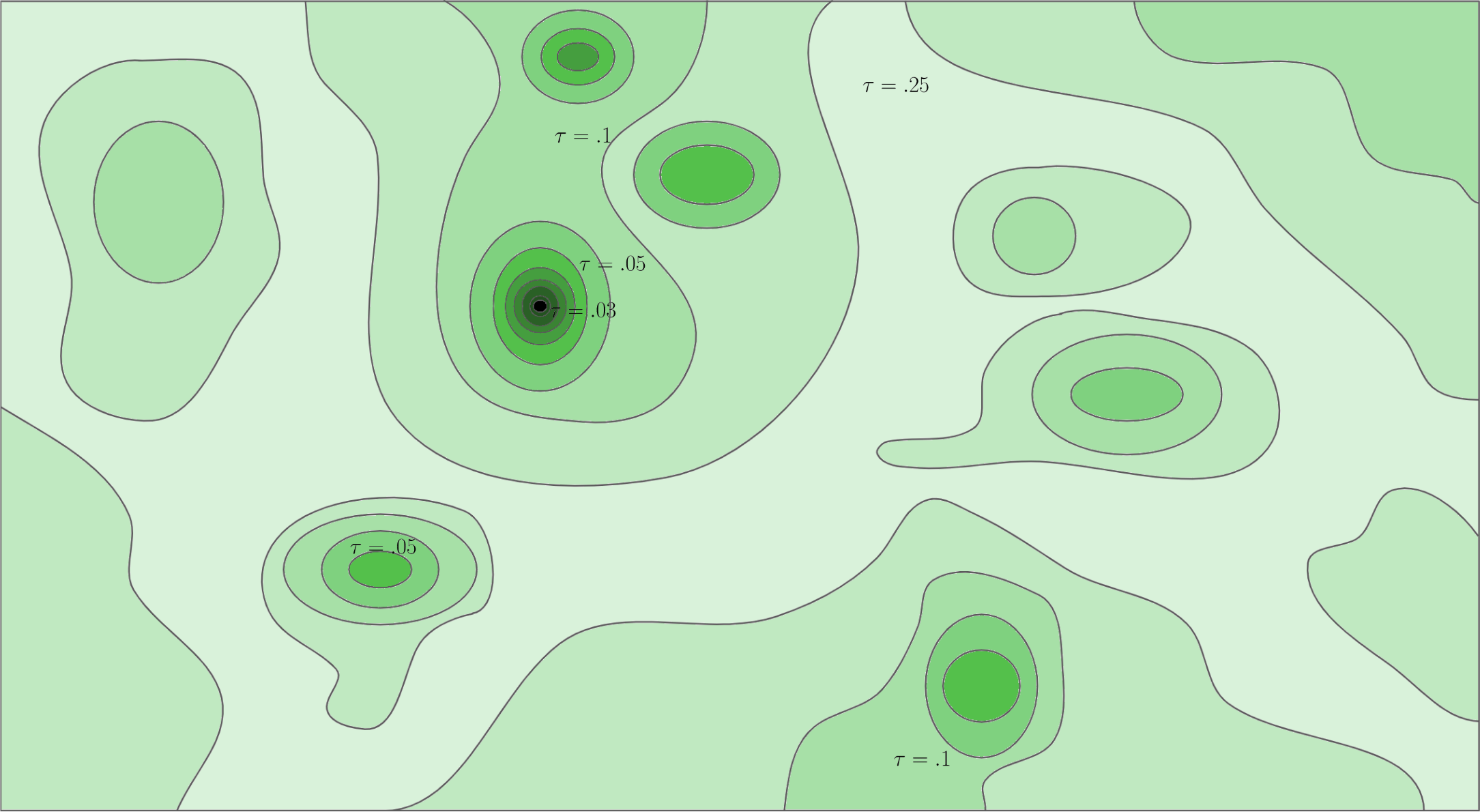}}
\caption{Non-convex solution space for Anchor Reconstruction. Darker contours imply a smaller reconstruction error.}
\label{icml-historical}
\end{center}
\end{figure}

\begin{algorithm}
  \caption{Sample $\xi' < \xi$
    \label{alg:packed-dna-hamming}}
  \begin{algorithmic}[1]
    \Require{Max Threshold $\xi$}
    \Function{Threshold-Sample}{$\xi$}
      \While{True}
        \State $\xi' \sim \mathcal{N}(\mu = \xi, \sigma = \xi / 6)$
        \If{$\xi' > 2\xi$ or $\xi' < 0$  }
        \State \textit{continue}
        \EndIf
        \If{$\xi' > \xi$}
        \State $\xi' \gets \xi - \xi'$
        \EndIf
        \State \Return $\xi'$
      \EndWhile
    \EndFunction
  \end{algorithmic}
\end{algorithm}

\begin{figure}[ht]
\begin{center}
\centerline{\includegraphics[width=\columnwidth]{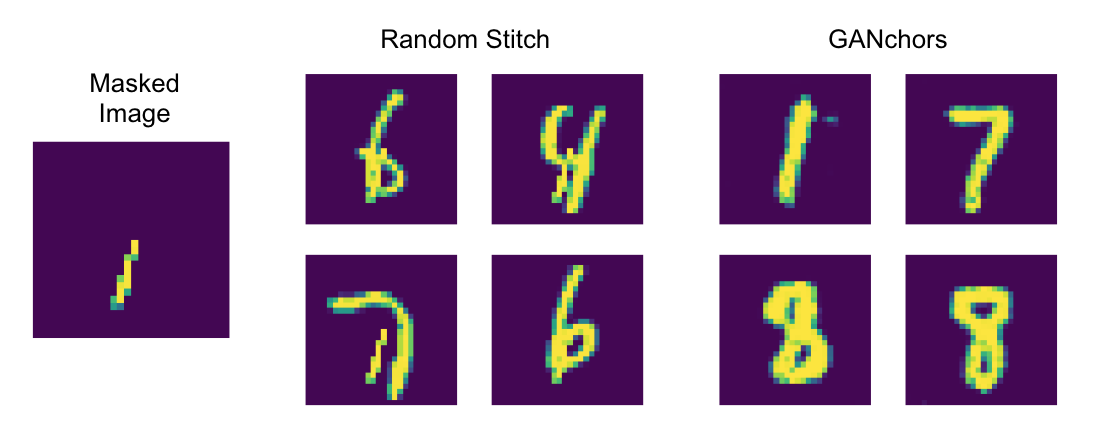}}
\caption{Comparison of Sampling using MNIST}
\label{icml-historical}
\end{center}
\end{figure}

\section{Speed Improvements}

When we implement Algorithm 1 to compute anchors, it quickly becomes intractable in that the \textit{KL-LUCB} algorithm to estimate the precision will require many samples. In following sections, we will cover improved methods that increase the computational efficiency of GANchors. 

\subsection{Batch Sampling}

One natural improvement to Algorithm 1 is to initialize a large batch of $z$ and jointly optimize each under the mean loss. This is easily done in deep learning libraries such as PyTorch \citep{paszke2017automatic}. To make this as efficient as possible, we apply additional modifications. First, we only update latent vectors $z$ if their reconstruction error of the anchor is above the desired threshold. If they are below the threshold, we instantly collect the reconstructions $G(z)$ and replace these latent vectors with new ones from the same initial distribution. 

Secondly, if applying Algorithm 2 to obtain several possible $\xi'$, a naive matching algorithm would suffer from unlucky initialization. To remedy this, we sort latent vectors $z$ by their anchor reconstruction loss, decreasing, and sort thresholds $\xi'$, increasing, and match based on the same indices. We chose to do this because it can be done easily using vectorized operations on tensors. This approach is summarized in Algorithm 3.

\begin{algorithm}
  \caption{Batch Sample $y\sim D(\cdot|A)$
    \label{alg:packed-dna-hamming}}
  \begin{algorithmic}[1]
    \Require{Trained GAN $G$, Thresholds $\{\xi'\}$, Anchor $A$, Target $\hat{x}$, Learning rate $\alpha$, Number of Samples $N$}
    \Function{Sample-Batch}{$G, \{\xi'\}, A, \hat{x}, \alpha, N$}
      \State Fill set $Z$ with $N$ samples from $z\sim\mathcal{N}(0,1)$
      \State $Y \gets \emptyset$
      \State $\Xi \gets \{\xi'\}$
      \While{$|Y| < N$}
        \State $\mathcal{L} \gets \{||\hat{x} - AG(z_i)||\}$
        \State Sort $\mathcal{L}$ increasing. Use indices to resort $Z$.
        \State Sort $\Xi$ decreasing
        \State Match $Z$ and $\Xi$
        
        \State $\hat{Z} \gets \{z_i \ s.t.||\hat{x} - AG(z_i)|| \leq \xi_i' \}$
        \State $Z \gets \{z_i\ s.t. \ ||\hat{x} - AG(z_i)|| > \xi_i'\} $
        
        \State $Y \gets Y \cup (1-A)G(\hat{Z}) + \hat{x}$
        \State Remove $\xi_i'$ used in $\hat{Z}$ from $\Xi$
      \EndWhile
      \State \Return{$Y$}
    \EndFunction
  \end{algorithmic}
\end{algorithm}

\subsection{Diverse Encoder}

When applying our algorithms to our datasets as seen in Section 5, we found that the speeds were still incomparable to the original method. To push this further, we make use of a technique used in \citep{zhu2016generative}. Their objective was to project an input image $x$ to the manifold of possible images generated on a trained GAN, by following a similar optimization to Algorithm 1, without any masking. They also noted that this is highly dependent on initializations and wanted to reduce time taken. As a solution, they incorporate another encoder network $P$ that maps the input image to some vector in the $d-$dimensional latent space of the GAN $G$. This encoder is optimized with the following:
\begin{equation}
    \min_{\theta_P} \mathbb{E}_{x}[ || x - G(P(x)) ||] 
\end{equation}
Essentially, their goal is to have some deterministic network to immediately project the input image to a good reconstruction that lies on the manifold of $G$. This provided a significant speedup to their setup; however, this is not immediately usable in our case. Our goal is to replace the initial sampling of our latent vectors $z$ with a similar network, but if they all begin at one point (since $P$ is deterministic), then we will no longer get diverse samples.

To solve this problem, we instead create a \textit{diverse encoder} $\hat{P}$, that outputs more than one vector $z$. Suppose that the latent space of $G$ is in $\mathbb{R}^{100}$. An encoder from \citep{zhu2016generative} would output a $100$-dimensional vector. Instead of doing this, $\hat{P}$ will output an $N \times 100$-dimensional vector where $N$ denotes the number of possibly encoded vectors desired from the network. So, if $N=8$, then it will output $8$ possible latent vectors, from a single input image $x$: $\{z_i\ i=1,...,8\}$ such that each $G(z_i)$ is a faithful reconstruction of $x$. However, we want to encourage diversity -- trivially, $\hat{P}$ can learn to output $N$ copies of the same vector, and we reach a similar issue. Thus we add in an additional regularization term that computes pairwise distances between the different encodings:
\begin{equation}
    \min_{\theta_{\hat{P}}} \mathbb{E}_{x}[ \sum_{z_i \in \hat{P}(x)}^N|| x - G(z_i) || - \lambda \sum_{i\neq j}^N || z_i - z_j || ]
\end{equation}

Notice that because we are trying to minimize (10), this would push the different outputs of $\hat{P}$ towards different locations in the latent space. Because this is unconstrained (since the network could just output very large and different vectors but not any good reconstructions), we apply a maximum $t$ on the norm of the $\lambda$ term. Numerically it is modified to the following:

\begin{equation}
    \min_{\theta_{\hat{P}}} \mathbb{E}_{x}[ \sum_{z_i \in \hat{P}(x)}^N|| x - G(z_i) || + \lambda \sum_{i\neq j}^N (\frac{|| z_i - z_j || - t}{t})^2 ]
\end{equation}

Lastly, because our sampling algorithm takes a masked image $\hat{x}$ as input, we train $\hat{P}$ to reconstruct masked images instead. The final version can be found in (12), where $A$ is a randomized segmentation map. To augment Algorithm 3, we replace the initialization of $z$ in Line 2 by first encoding the target $\hat{x}$ and then adding random noise to help induce more diversity.

\begin{equation}
    \min_{\theta_{\hat{P}}} \mathbb{E}_{x}[ \sum_{z_i \in \hat{P}(x)}^N|| Ax - AG(z_i) || + \lambda \sum_{i\neq j}^N (\frac{|| z_i - z_j || - t}{t})^2 ]
\end{equation}

\section{Experiments and Results}
\subsection{MNIST}

To initially validate our various algorithms during testing, we opted to use MNIST \citep{lecun-mnisthandwrittendigit-2010}. We first trained a small convolutional neural network to predict the class and achieved a 99\% accuracy on the test set. We also use a pretrained DCGAN implementation \citep{CSINVA} for all of our testing. To run explanations using our algorithm, we set a max anchor MSE threshold $\xi = 0.05$. To obtain samples for speed and precision analysis, we took a single test image from each class and ran three trials on each to compare the different methods in Section 4 as well as their baseline method. We limited testing to this scenario due to the large complexity of the naive methods. For segmentations, we use the quickshift algorithm from \textit{scikit-image}. We wanted each example to produce exactly 15 segments each, so we modify the max distance parameter through a binary search to obtain the desired number of segments.

\begin{table}[H]
\begin{center}
    
  \label{tab:time_mnist}
  \begin{tabular}{l|c}
    \toprule
    \textbf{Sampling Method} & \textbf{Time (min.)}\\
    \hline
    Random Stitching & 0.08\\
    \hline
    Batch & 34.87\\
    Batch + Batch Norm & 4.46\\
    \hline
    Diverse Encoder & 1.62\\
  \bottomrule
\end{tabular}
\caption{Average Explanation Time}
\end{center}
\end{table}

For batch processing, we use a batch size of 64 and apply random restarts every 1000 iterations to help escape suboptimal local minima on the manifold. We use the Adam optimizer to run gradient descent. During experiments, we noticed that batched processing took much longer than expected; upon inspection of the GAN's architecture, we found batch normalization layers \citep{Ioffe:2015:BNA:3045118.3045167}. We hypothesize that this created a significant bottleneck for our algorithm. 

Batch normalization aims to solve the problem of internal covariate shift, or unscaled input distributions within each layer of the network. The hypothesis is that network performance, originally on classification tasks, can be improved significantly if inputs to each layer can be normalized, much like how input features are normalized during pre-processing stages. This has also extended to generative networks, as it can lead to more stable training. During training, the layers collects the minibatch mean ($\mu_{\mathcal{B}}$) and variance ($\sigma_{\mathcal{B}}^2$) of its inputs and compute the new output $y$ with the following:

\begin{align*}
    \hat{x}_i &\gets \frac{x - \mu_{\mathcal{B}}}{\sqrt{\sigma^2_{\mathcal{B}} + \epsilon}} \\
    y_i &\gets \gamma \hat{x}_i + \beta\\
\end{align*}

\begingroup\vspace*{-\baselineskip}
\captionof{figure}{Batch Normalization Layer Operations}
\vspace*{\baselineskip}\endgroup

The learned parameters are $\gamma$ and $\beta$. The issue arises during the layer's behavior during training and testing mode. In training mode, it uses $\mu_{\mathcal{B}}$ and $\sigma^2_{\mathcal{B}}$, but \textit{also} computes a running mean and variance, $\mu_{\mathcal{D}}$ and $\sigma^2_{\mathcal{D}}$, respectively. In testing mode, it uses these estimates $\mu_{\mathcal{D}}$ and $\sigma^2_{\mathcal{D}}$ to normalize, with the same scheme as in Figure 5. This is an issue, because these learned estimates from the pretrained GAN do not reflect the new batches during our sampling. The original training assumed an input distribution of $\mathcal{N}(0,1)$, whereas our input distribution is constantly changing -- it is only optimized so that every output from the GAN achieve an MSE reconstruction threshold of the desired anchor. When only using training mode, our time significantly decreases, as seen in Table 1. 

Lastly, we implemented a diverse encoder for the MNIST dataset. We first run the quickshift algorithm on the dataset to obtain segmentations beforehand - this would speed up training time as they can be expensive to compute in real time. Then, we create a small convolutional neural network that takes a \textit{masked image} as input and then outputs 8 possible encodings, by outputting an 800 dimensional vector, since the latent space of the GAN is in $\mathbb{R}^{100}$. To train this we use the loss function in Equation (8) of Section 4 with $\lambda = 1$, and $t = 10$. We also apply an upperbound on the norm of pairwise errors to be 10. The input segmentations are randomized, by selecting each segment label with a probability of 0.5. We apply early stopping after a single iteration through the training set. We show examples of both good and bad results in Figure 6 and 7.

\begin{figure}[ht]
\begin{center}
\centerline{\includegraphics[width=\columnwidth]{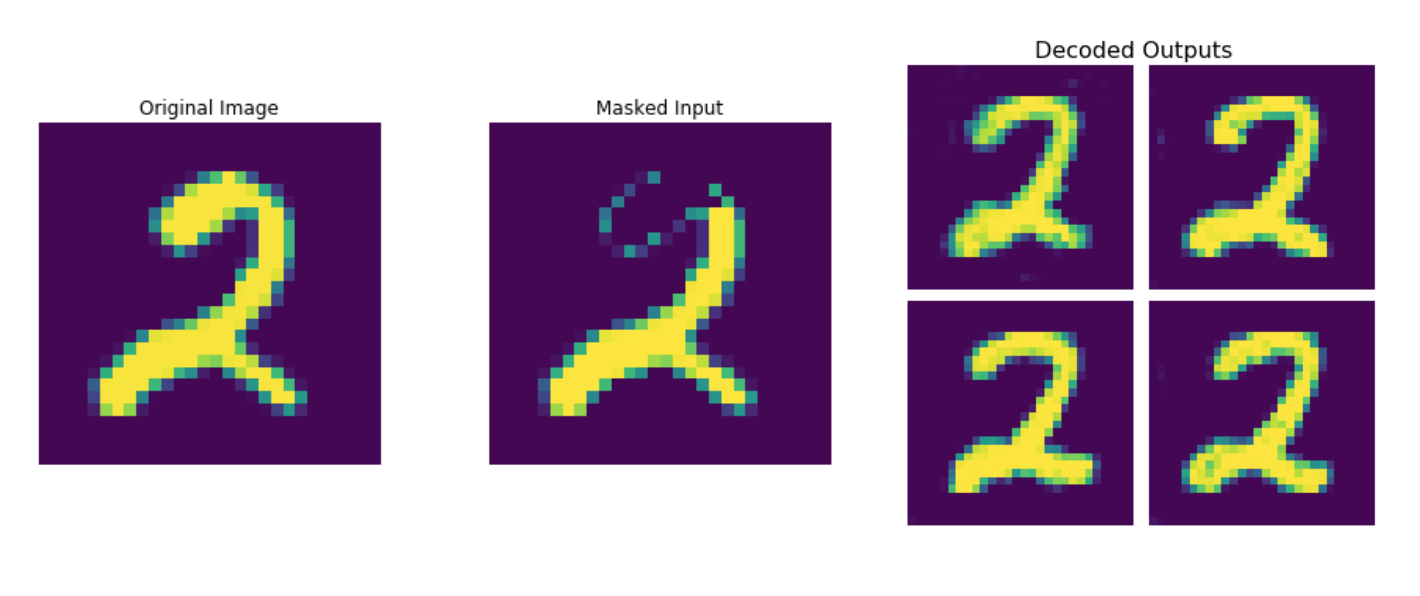}}
\caption{Good Diverse Encoder Outputs }
\label{icml-historical}
\end{center}
\end{figure}

\begin{figure}[ht]
\begin{center}
\centerline{\includegraphics[width=\columnwidth]{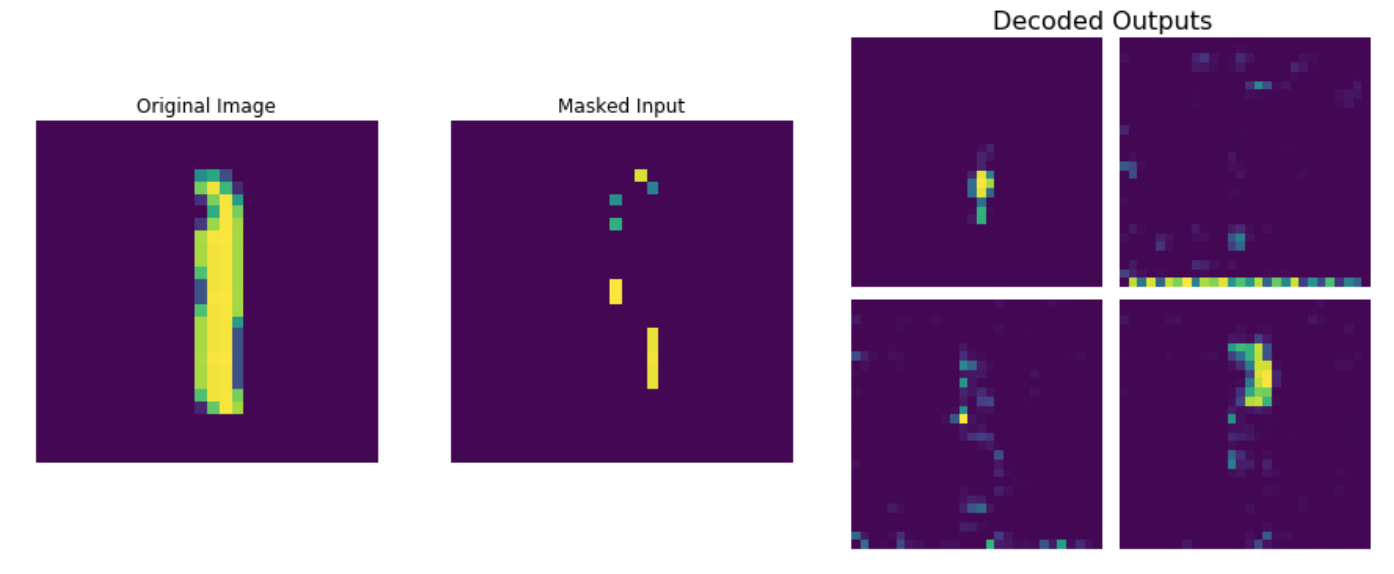}}
\caption{Bad Diverse Encoder Outputs }
\label{icml-historical}
\end{center}
\end{figure}

First note that this diverse encoder is only trying to produce encoded vectors that when decoded and masked, match well the the masked input. The reason we distinguish between good/bad in these figures are the whole images produced by the network. In Figure 6, it actually produces realistic two's similar to the original image, but Figure 7 shows noisy and nonsensical images. This is most likely due to a small amount of disconnected signals that are used as inputs to the encoder. Nevertheless, we are only looking to use these encoded vectors as \textit{starting areas} for optimization, so they do not have to be perfect at reconstruction. 

To test the effectiveness of these methods, we first measure the average time to produce a single explanation with different sampling methods for $\mathcal{D}(z | A)$. These can be found in Table 1. We see that improvement to sampling changes the average time by an order of magnitude, however cannot reach the times produced by the original random sampling method. We believe that if the encoder was more finely trained, faster times can be achieved, but that tradeoff is acceptable to a certain extent for more realistic sampling.

Additionally, we measure the precision of our explanations on the same set as another metric of comparison. These can be found in Table 2. Immediately we can see that our explanations achieve a higher precision, and smaller sizes of anchors. We hypothesize that this happens due to the inherent structure of natural images, especially those that are learned by a trained GAN. Given a subset of pixels that must be reconstructed, there exist a smaller amount of possible digits. For example, in Figure 6, we can see that the structure of the masked input contains the curves one would only see in handwritten 2s. There exist correlations between the grouping of pixels that actually provide more information for the classification, since the GAN would only create perturbation that look like 2s. Thus, as the anchor size increases, the classifier sees a more and more restricted perturbation distribution. We assert that this is expected behavior, as we are now only testing the classifier on true digits, rather than a random mixture of different parts. This however weakens the explanation, as the segments shown now describe \textit{pixels that result in the same prediction, or are correlated to other pixels that result in the same prediction}. Further work must still be done to strengthen these conclusions. Nevertheless, the explanations are more faithful to the true distribution of data, as the explanations from the original random stitching method can also be a result of correlations between outlier signals of the random images. We have also included side by side comparisons in Figure 8.

\begin{table}[H]
  \begin{center}
  \label{tab:mnist_prec}
  \begin{tabular}{l|c | c}
    \toprule
    \textbf{Sampling Method} & \textbf{Precision} & \textbf{Size of Anchor}\\
    \hline
    Random Stitching & 0.982 & 4.2\\
    \hline
    Batch + BN & 0.987 & 1.87\\
    \hline
    Diverse Encoder & 0.993 & 1.63\\
  \bottomrule
\end{tabular}
\caption{MNIST Anchor Precisions. We omit non Batch Norm results as samples were initialized the same way. Size of Anchor defined as the average number of segments. }
\end{center}
\end{table}

\begin{figure}[h]
\vskip 0.2in
\begin{center}
\centerline{\includegraphics[width=0.6\columnwidth]{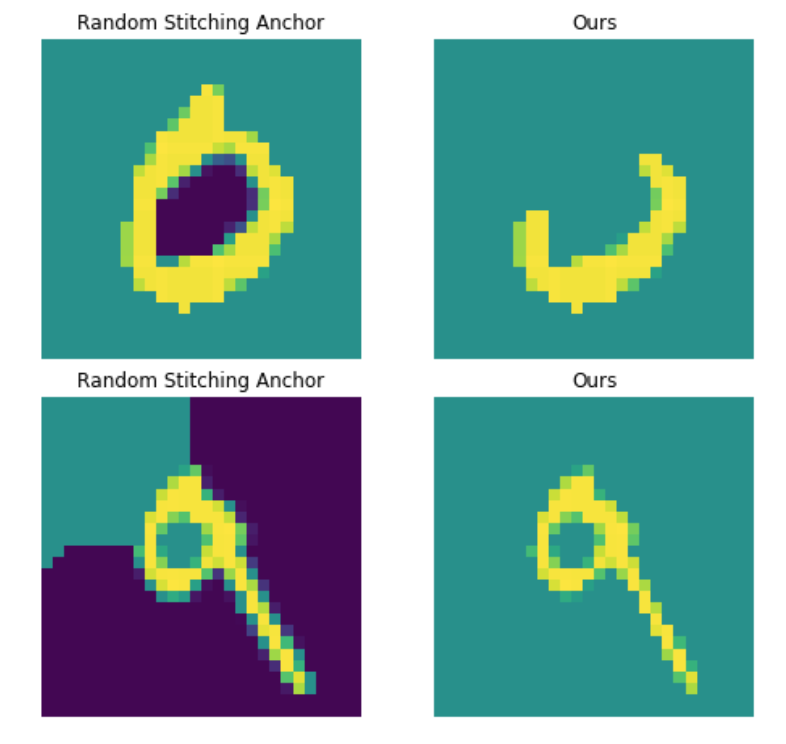}}
\caption{Computed Anchor Comparisons \textbf{\underline{Left}:} Original Random Stitching Sampling. \textbf{\underline{Right}}: Sampling from a trained GAN}
\label{icml-historical}
\end{center}
\vskip -0.2in
\end{figure}

\subsection{CelebA-HQ}
While the original authors' explored anchors on ImageNet, state of the art GANs for the dataset were mainly class-condition, including BigGAN. It proved to be a challenge to generate samples below a desired threshold $\xi$ while selecting random classes. Due to this, we opted to use CelebA-HQ \citep{celebahq}, a dataset of 1024x1024x3 images of celebrity faces created by passing CelebA through a super-resolution network. The generative model associated with this dataset is ProGAN \citep{karras2018progressive}, a progressively grown generator which outputs high resolution, realistic 1024x1024x3 images of faces. For the purposes of computation time and memory, we scaled down both the dataset and generator output to 256x256x3. 

For the purposes of testing, we trained a convolutional neural network using ResNet18 to predict whether the image was smiling or not. The classifier we attempted to explain achieved a validation accuracy of 93\%. The units for CelebA anchors were segments achieved using SLIC (numsegments = 10, compactness = 20). Finally, we had to qualitatively select an anchor MSE threshold at which to collect the samples based on running time, quality of the match, and variation of results. After testing in the range of [0.01, .15], we found that .075 would be sufficient for our experiments. The results of this threshold testing can be seen in Figure 9. The CelebA experiments were ran on a single GTX1080 with an i5-4790k at 3.5Ghz. 

The anchors our method found were consistent with intuition in that the anchor was the lower face, mouth, or combination of the two. Naturally, these are important segments for smiling and thus result in a valid explanation of the image. All of the anchors seen in Figure 10 generated by our algorithm achieved a precision of 1.0. 

\begin{figure}[ht]
\vskip 0.2in
\begin{center}
\centerline{\includegraphics[width=\columnwidth]{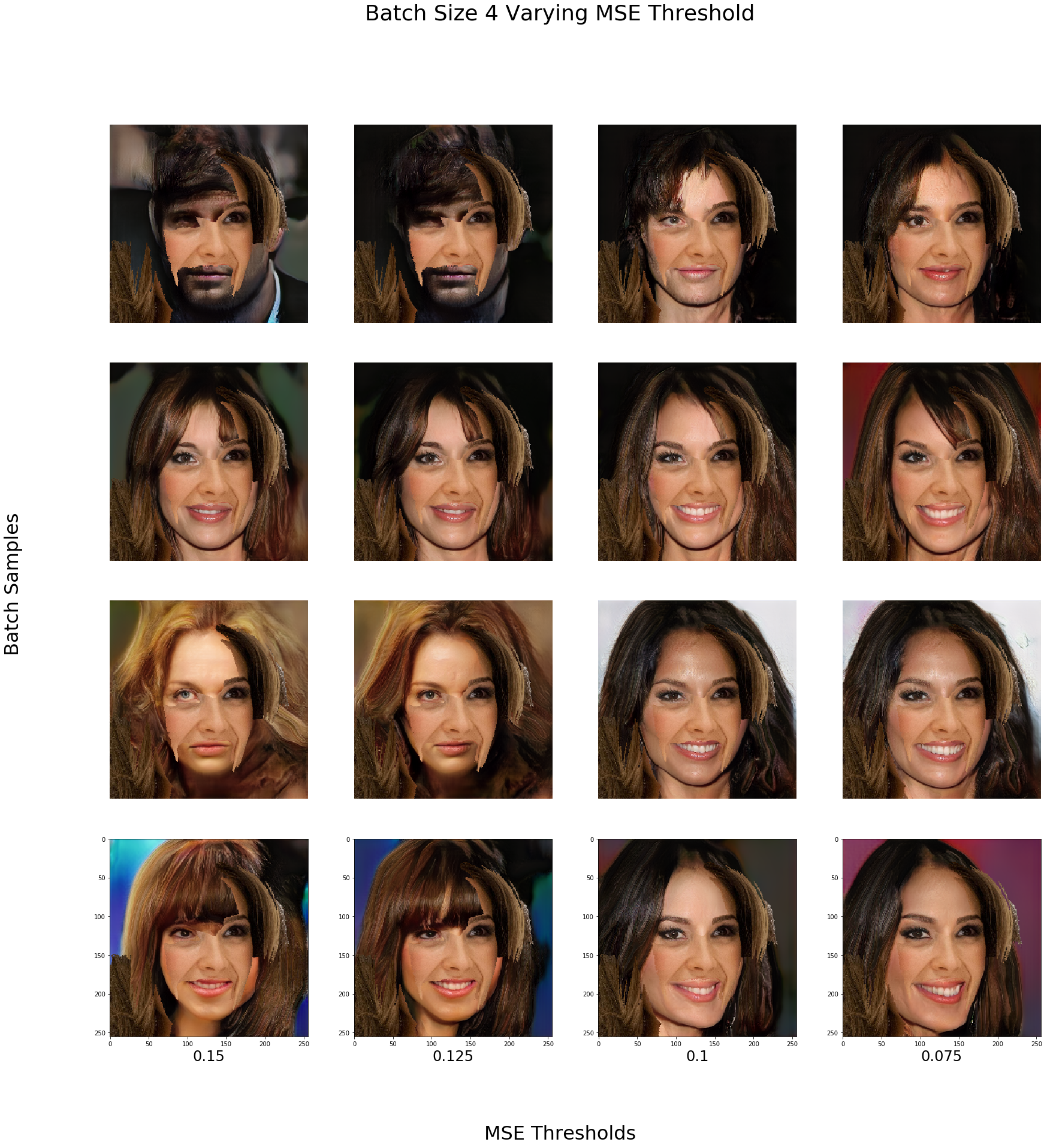}}
\caption{Batch samples generated at various MSEs. As the threshold decreases, the samples have lower variance, but higher running time. Additionally for anchor purposes, a baseline variance is necessary to actually test the potential anchor. We found .075 balanced the these tradeoffs appropriately. }
\label{icml-historical}
\end{center}
\vskip -0.2in
\end{figure}

\begin{figure}[h]
\vskip 0.2in
\begin{center}
\centerline{\includegraphics[width=\columnwidth]{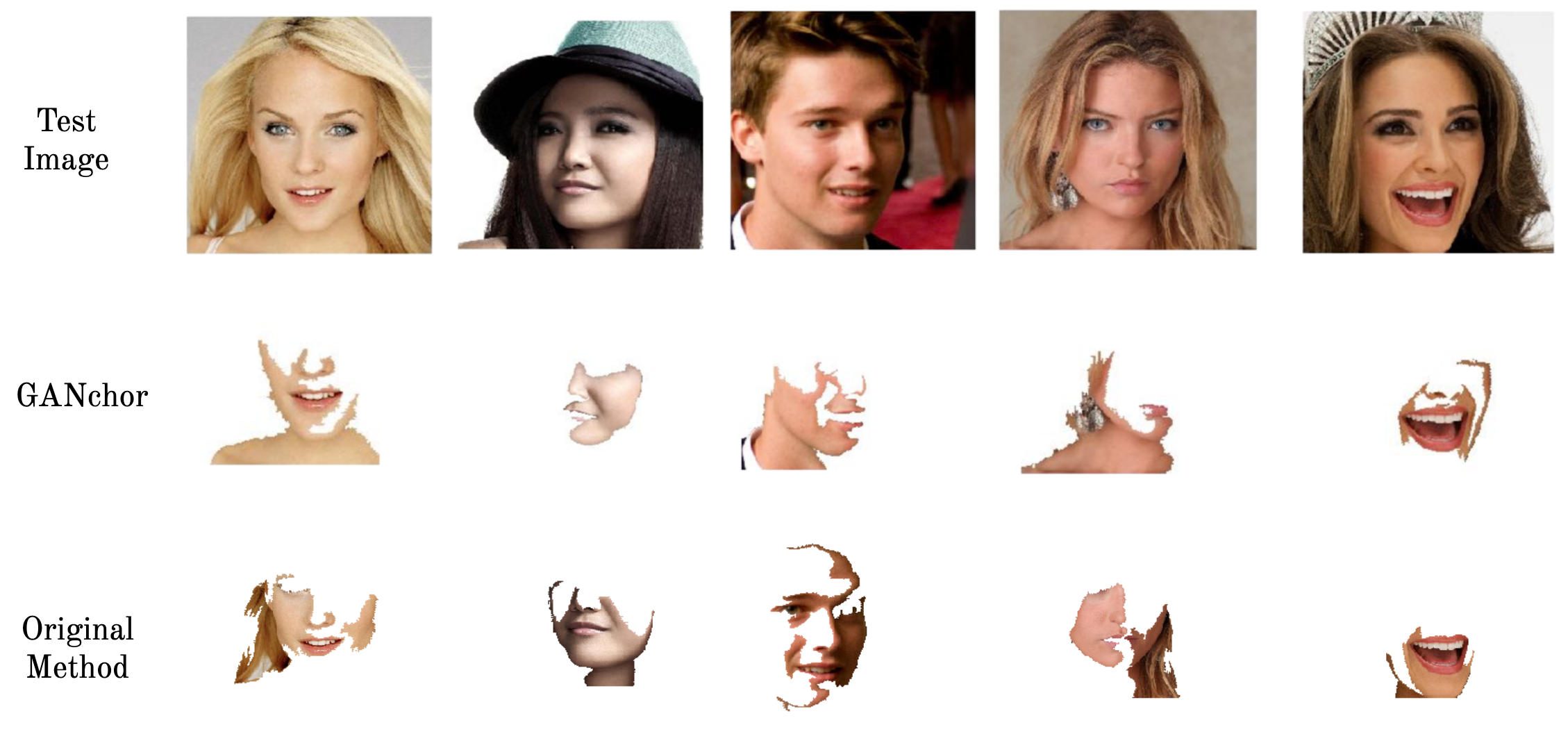}}
\caption{Anchors for the prediction of smiling or not smiling. For most cases, the anchor involved the lower face and mouth, intuitively sufficient for smiles. All anchors seen above achieved a precision of 1.0. Compared with the bottom row from the original method, our anchors tend to be smaller with better precision.}
\label{icml-historical}
\end{center}
\vskip -0.2in
\end{figure}

We also try to train a diverse encoder on the CelebAHQ dataset. To do this, we obtain masks from CelebAHQ-Mask \citep{celebahqmask} and create an encoder that outputs $N=2$ encodings. This is limited due to the large memory consumption needed to use ProGAN. We trained on random segmentations, much like in MNIST, with a batch size of 8. To stabilize training, we found that we also had to apply $L2$ regularization on the encoder's outputs, as they tend to diverge towards large vectors. We use $\lambda = 1e$-$3$ and L2 regularization parameter of $1e$-$6$. We have included results of an example encoding and patched up results in Figure 11. Unlike MNIST, our encoder was unable to achieve both decent reconstructions as well as diversity. The reconstructions themselves look very blurry, as if they are average images in the dataset; it proved difficult to train due to the sheer size of the input images. In addition, when we used these as starting points for our batch search, this resulted in all very similar images, with less quality than compared to Figure 9. For this reason, we stuck to randomly initializing from $\mathcal{N}(0,1)$.

\begin{figure}[ht]
\vskip 0.2in
\begin{center}
\centerline{\includegraphics[width=\columnwidth]{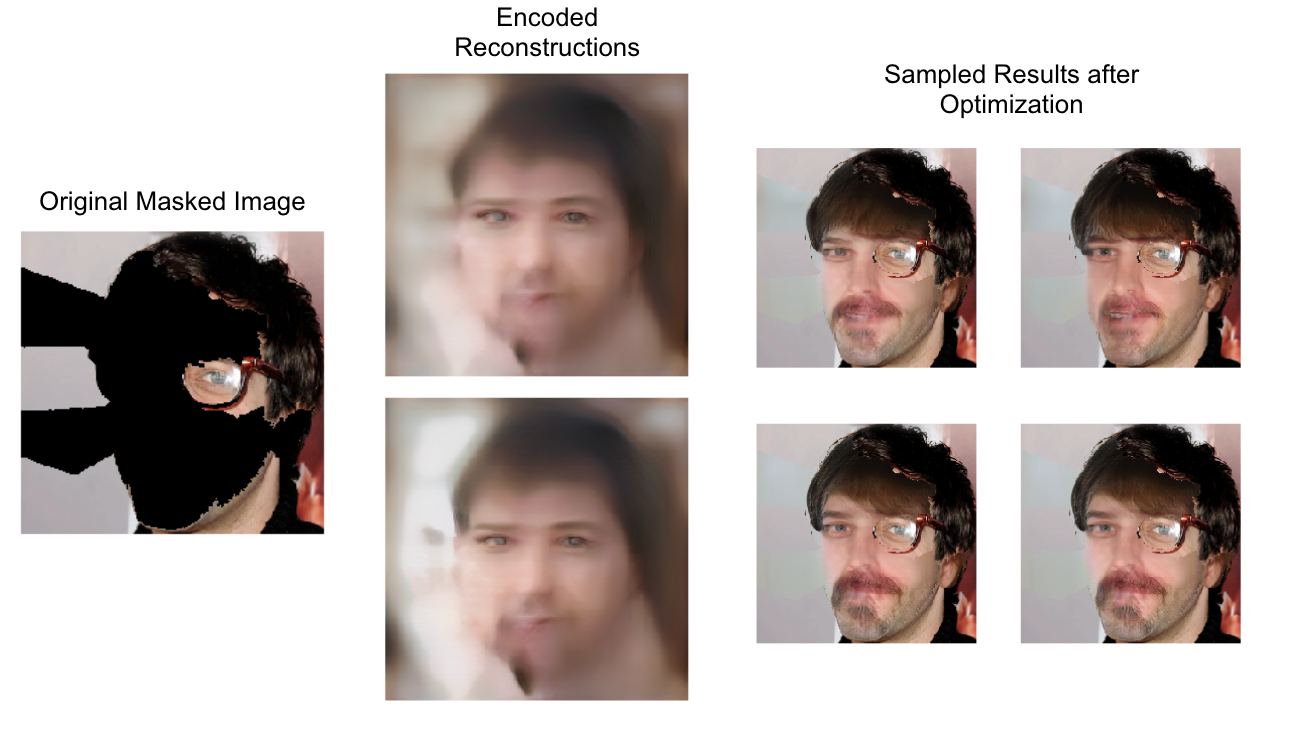}}
\caption{Diverse Encoder on CelebA-HQ. \textbf{\underline{Left}}: Original Masked Image \textbf{\underline{Middle}}: Reconstructions from Diverse Encoder Outputs \textbf{\underline{Right}}: Results from applying our Sampling Algorithms}
\label{icml-historical}
\end{center}
\vskip -0.2in
\end{figure}

\section{Discussion}

\subsection{Limitations}
A considerable drawback of this approach is the computational complexity of having to run a GAN. Even though we were able to dramatically increase the average speed of generating anchors for images from hours to a few minutes, the time to train a GAN as well as sample enough coverage samples for a robust anchor is still considerably high,  for example if one wanted to generate explanations for a large dataset. Furthermore, running ProGAN with a batch size of 4 inside our anchor sample generation algorithm required at least 8GB of GPU memory on a single GTX1080. The high memory consumption limits the speedup gained through batch, and could potentially be optimized for a larger speedup. This bottleneck also makes it difficult to train a successful diverse encoder on large datasets, but we hope that the idea can be explored more.

Our work also relied heavily on qualitatively searching for optimal hyperparameters for generating realistic images. We considered various boundary error methods to measure how well a sampled background fit the anchor, but in cases with a clean segmentation cut there would be higher acceptable boundary differences than an anchor in the center of a homogeneous area. Without making considerable assumptions, we were unable to define a contextual realism metric of how well an anchor maps onto a sample without extending the anchor outside of its bounds or using information from the original image not found in the anchor. 

Finally, our work assumes that an unconditional GAN exists for the data distribution that the classifier is trained on. Thus, the challenges of training a GAN are transferred to our work. This difficulty can be relieved through sharing of models and public access to datasets and generators.

\subsection{Future Work}
While our project was limited to the image domain, the use of GANchors could be applied to other domains, be that text, tabular or time series. As GANs model a mapping from a latent space to the underlying data distribution, they could theoretically be used to sample from any distribution in most domains. By incorporating GANs into the anchor pipeline, this technique could be increasingly model-agnostic, simply accepting a generator and explainable feature extractor specific to the model/domain. 
In addition, we would also like to supplement our hypotheses of more realistic perturbations by developing a new metric that could reflect the similarity to a reference set. Something along the lines of a BLEU score but for images is what we have in mind.


\bibliography{example_paper}
\bibliographystyle{icml2019}





\end{document}